# Extraction of Typical Operating Scenarios of New Power System Based on Deep Time Series Aggregation

Zhaoyang Qu [1,2], Zhenming Zhang [1,*], Nan Qu [3], Yuguang Zhou [4], Yang Li [1], Tao Jiang [1], Min Li [1], Chao Long [5]
1 School of Electrical Engineering, Northeast Electric Power University, Jilin 132012, China
2 Jilin Province Technology Research Center of Power Big Data Intelligent Processing, Jilin 132012, China
3 State Grid Jiangsu Electric Power Co., Ltd. Nanjing Power Supply Branch, Nanjing 210008, China
4 State Grid Jilin Electric Power Co., Ltd., Changchun 130022, China
5 Department of Electrical Engineering and Electronics, University of Liverpool, Liverpool L69 3GJ, UK

**Abstract**: Extracting typical operational scenarios is essential for making flexible decisions in the dispatch of a new power system. This study proposed a novel deep time series aggregation scheme (DTSAs) to generate typical operational scenarios, considering the large amount of historical operational snapshot data. Specifically, DTSAs analyze the intrinsic mechanisms of different scheduling operational scenario switching to mathematically represent typical operational scenarios. A gramian angular summation field (GASF) based operational scenario image encoder was designed to convert operational scenario sequences into high-dimensional spaces. This enables DTSAs to fully capture the spatiotemporal characteristics of new power systems using deep feature iterative aggregation models. The encoder also facilitates the generation of typical operational scenarios that conform to historical data distributions while ensuring the integrity of grid operational snapshots. Case studies demonstrate that the proposed method extracted new fine-grained power system dispatch schemes and outperformed the latest high-dimensional feature-screening methods. In addition, experiments with different new energy access ratios were conducted to verify the robustness of the proposed method. DTSAs enables dispatchers to master the operation experience of the power system in advance, and actively respond to the dynamic changes of the operation scenarios under the high access rate of new energy.

**Keywords**: new power system; high proportion of new energy; power system operation mode; operation scenario image encoder; convolutional neural networks; deep time series aggregation.

## 1 Introduction

In the new power system (NPS) with new energy as the main body, wind power and photovoltaic have become the main power sources of the national power system, and the flexible loads on the demand side will actively participate in the operation control of the power grid [1-2]. In this context, the power supply output fluctuation cycle is shortened to the minute level, and load changes are not just based on the forecast curve, but will be guided by price fluctuations in the power market. That is, changes in power system operating scenarios are accelerating [3]. In the traditional power dispatch mode, the operating scenario change cycle is 15-30 minutes. The contradiction between the two time scales reveals that the current level of granularity of the operation scenario cannot adapt to the rapidly changing operating form of the new power system. In addition, the ability to respond to uncertainties from multiple sources, such as new energy sources and flexible loads, determines the future proportion of renewable energy integration and energy consumption rate [4]. Therefore, mastering the changing patterns and key features of scheduling and operational data on a massive scale, as well as exploring methods for extracting typical operating scenarios for new power system, are strategically significant aspects [5].

### *1.1 Related works*

#### 1.1.1 Artificial extraction of power system operating scenarios

The extraction of traditional power system operating scenarios mainly relies on manual operating experience. Based on manual analysis of historical operation conditions, the characteristics of power grid operation are summarized, and a power grid operation mode report and stable operation procedures are compiled [6].This method can still be applied to traditional power systems with concentrated and relatively single operating scenarios. However, as the penetration rate of new energy continues to increase, system operating scenarios are characterized by high randomness and diffuse distribution [7].

It is difficult for artificial knowledge to take into account the changes in operating scenario data distribution brought about by new energy access, resulting in insufficient applicability of the extracted operating scenarios under NPS morphology.

**1.1.2 Extraction of operating scenarios based on clustering algorithm**

Hou et al. [8] mentioned that power system operation scenario identification is a typical big data analysis problem. Focusing on the wide-area measurement data of system operation, PCA and k-means++ algorithms were used to conduct a preliminary exploration of operating mode clustering. Wang et al. [9] combined traditional clustering, feature extraction and vine copulas methods to propose a data-driven system scheduling operation scenario generation architecture under renewable energy. Niu et al. [10] used K-means clustering method to extract extreme dispatch scenarios from random scenarios of power grid operation for analysis, which is of great significance to the optimized operation of power systems with the participation of wind/solar energy. Tso et al. [11] proposed clustering massive power grid historical operation snapshots based on hierarchical clustering algorithm to obtain stable operation rules and provide advanced safety warnings for the system. Ren et al. [12] constructed a feature quantity library that represents system operating snapshots, used a decision tree model to screen the feature quantities, and finally used a data-driven method to perform similarity clustering on historical operating snapshots.

This type of method follows the mathematical paradigm of feature variable screening + traditional clustering algorithm, and there is no interaction between the two. This not only leads to the destruction of the integrity of historical dispatch experience, but also makes it impossible to perform cluster analysis on long-term system operation snapshots[13].

**1.1.3 Operation scenario generation based on deep learning**

With the improvement of computing power and the rapid development of artificial intelligence technology represented by deep learning [14], it has gradually been applied to the generation of operating scenarios of power systems. Deep learning models applied to sample generation problems include variational auto-encoder (VAE) [15-16], deep belief network (DBN) [17], and generative adversarial networks (GAN)[18] etc. Yuan et al. [19] proposed a method based on multi-objective GAN for power grid scene generation considering the uncertainty of wind power. However, the accuracy and diversity goals of the generated scenes require careful design, otherwise the generated scenes will be affected. performance. Karras et al. [20] controlled the latent space inside the neural network by improving the generator structure, thereby reducing sample dependence. Wang et al. [21] proposed a power system operating scenario generation method based on graph representation learning and feature guidance, using graph representation learning to enhance the expression of features. Experiments proved that the model can generate operating scenarios while improving the generation efficiency of specified feature operating scenarios.

Although this type of method can learn the distribution of historical operating scenario data to generate new scenarios, in multi-uncertain environments with numerous scheduling targets (e.g., traditional generators, renewable energy units, energy storage, and adjustable loads), the traditional numerical sample expression form greatly limits the performance of deep learning algorithms.

***1.2 Motivation and contributions of this paper***

In view of the above reasons, this work focuses on the evolution process of NPS and discusses the refined analysis and extraction schemes of typical operating scenarios under different new energy penetration rates. We solve the problem of extracting typical operating scenarios of NPS by enhancing the form of data expression and realizing the interaction of feature extraction and clustering. The former converts one-dimensional operating snapshot samples into three-dimensional feature image samples. The interaction between feature extraction and clustering means that both use a unified loss function, and the clustering results simultaneously guide the parameter updates of the feature neural network and clustering algorithm.

The main contributions are as follows:

- A new approach to power system operation scenario recognition has been developed, using a novel power system operation scenario image encoder based on Gramian Angular Summation Field (GASF). This approach generates image datasets of power-grid operation scenarios while preserving the experience of the dispatchers.
- A novel neural network structure, called VGG13_GMP, was constructed for power system operation scenario image feature extraction using the first 13 convolutional layers of Visual Geometry Group 16 (VGG16) and one global max pooling layer (GMP). This approach provides a new perspective on the application of deep clustering methods for information extraction from grid operation snapshots.
- Further, a new power system typical operation scenario extraction scheme based on deep time-series aggregation (DTSAs) was proposed. This method iteratively performs DTSAs to generate typical operational scenarios that match the historical data distribution, while adapting to the increasing penetration rate of new energy.

### *1.3 Roadmap of this paper*

The remainder of this paper is organized as follows. Section 2 presents the mathematically characterization of the new power system operation scenarios considering the spatiotemporal integrity. Section 3 discusses the design of the GASF-based power system operation scenario image encoder. Thereafter, a typical operating scenario extraction model based on DTSAs is outlined in Section 4. Section 5 presents the results of the application of the proposed method to real-world datasets and verifies its advantages. Finally, Section 6 concludes the paper.

| **Nomenclature** | | | |
|---|---|---|---|
| *Sets* | | $n_{out}$ | the pixel value of the feature map after the convolution calculation |
| $S$ | set of operating snapshots for a continuous period of time | $p_{in}$ | the size of the input feature map before pooling calculation |
| $s_i$ | set of element power in the snapshot at time $i$ | $f_{size}$ | the size of the filter |
| $OM_i$ | the $i$ th typical operating scenario | $sd_{pool}$ | the moving step size of the filter |
| $\tilde{s}_i$ | the normalized sequence set of snapshots at time $i$ | $p_{out}$ | the size of the output feature map after pooling calculation |
| *Parameters* | | $w_k$ | the weight of the $k$ th Gaussian component |
| $g$ , $r$ | total number of traditional units or new energy units in operation | *Variables* | |
| $l$ | total number of load nodes | $G_{i,j}$ , $R_{i,j}$ | a variable for the power output of the $j$ th traditional unit or new energy unit in snapshot $i$ (MW) |
| $m$ | total number of traditional power generation, new energy, and load nodes in the regional power grid | $L_{i,j}$ | a variable for the power of the $j$ th load node in snapshot $i$ (MW) |
| $\theta_j$ | polar angle, the numerical relationship between data points | $GM$ | Gram matrix of a single operating snapshot |
| $r_j$ | polar axis, the spatial relationship between different data points | *Other* | |
| $n_{in}$ | the input pixel values before the convolution calculation | $L_{net}$ | feature extraction neural network loss in abstract form |
| $pd$ | the padding size at the edges of the convolutional layer | $L_{clustering}$ | clustering loss in abstract form |
| $sd_{conv}$ | the moving step size of the convolutional kernel | $\lambda_1$ , $\lambda_2$ | deep clustering model loss component weight |
| $k_{size}$ | the size of the convolutional kernel | | |

## 2 Mathematical representation of operating scenarios

As it is known, a power system comprises various components such as generators, transformers, transmission lines, and loads, all of which have distinct operating states. For instance, the generator can be in an on/off state, and the output size during operation can be considered a different operating state. Further, the different transmission power levels of the transmission lines determine the operating states of their components. In addition, the load undergoes state transitions with change in the demand. The "quality" of dispatching operations is heavily reliant on the dispatcher's "information processing ability" to coordinate the specific component states of each link based on resource status and demand analysis [22]. However, the combination of different component states is not arbitrary and must strictly follow the electrical constraints for each component, optimal flow of the network, and the objective function of the system.

Therefore, different combinations of operating states of the components at each link in the power system constitute different operating scenarios $s_i$, as shown below.

$$S=\begin{bmatrix} G_{1,1} & \cdots & G_{1,g} & R_{1,1} & \cdots & R_{1,r} & L_{1,1} & \cdots & L_{1,l} \\ G_{2,1} & \cdots & G_{2,g} & R_{2,1} & \cdots & R_{2,r} & L_{2,1} & \cdots & L_{2,l} \\ \vdots & \ddots & \vdots & \vdots & \ddots & \vdots & \vdots & \ddots & \vdots \\ G_{t,1} & \cdots & G_{t,g} & R_{t,1} & \cdots & R_{t,r} & L_{t,1} & \cdots & L_{t,l} \end{bmatrix} \tag{1}$$

where $S=(s_1,s_2,\cdots,s_t)^T$ indicates that there is a timing relationship between consecutive operating snapshots.

In traditional power systems, the power supply is mainly provided by thermal and hydroelectric power generation. The power flow is unidirectional within the area, and the distributed and centralized loads, after being equivalent to high voltage levels, exhibit consistent characteristics. Here, the load quantity maintains the same order of magnitude as the generators. To ensure the safety and stability of the power grid, this type of power system often employs a comprehensive model based on all components in the power grid to construct simultaneous mathematical equations. Moreover, its typical operating scenarios and dynamic response strategies are "exhaustive," thus, the system control center formulates different main dispatching modes at different time scales based on the system operation conditions of the previous year [23].

However, in the new power system, the number of participants in the power grid following the integration of a high proportion of new energy and flexible loads is massive. Moreover, the interest and responsibility entities are relatively independent, and the equivalent modeling idea has certain limitations. To achieve carbon reduction and low carbon emissions, it is essential to focus on the individual characteristics and interests of the generators and loads. Simultaneously, the complex and variable output of new energy during the operation process of the new power system, as well as the uncertainty on both ends of the source-load, result in a "combination explosion" of power grid operating scenarios. More importantly, traditional dispatching models focus on the extreme operating conditions of the power grid, and dispatchers only adhere to the safety bottom line when adjusting routine operating scenarios. This renders it challenging to dynamically respond to low-carbon goals. In summary, to form a refined scheduling and control plan, it is required to precisely divide the typical operating scenarios of the new power system in the temporal and spatial dimensions. The temporal dimension considers the time control ability of source-side output fluctuations and flexible loads. The spatial dimension considers the required resolution of generators and loads. By precisely dividing the typical operating scenarios of the new power system, we can develop more effective scheduling and control plans. This can satisfy the operational requirements of the "high proportion of renewable energy and power electronization" new power grid. Based on Equation (1), this study realizes a mathematical representation of the typical operating scenarios of a new power system considering spatiotemporal integrity.

$$OM_i=\{G^*_{i,1},\cdots,G^*_{i,g},R^*_{i,1},\cdots,R^*_{i,r},L^*_{i,1},\cdots,L^*_{i,l}\} \tag{2}$$

Faster support for changes in power grid operation modes, higher integration of new energy sources, and more complete demand-side response can be achieved by improving the ability to extract typical operating scenarios.

## 3 Operating scenario image encoder

Dispatchers have found that relying solely on mathematical models and automation solutions to ensure the smooth operation of a power system is challenging in actual practice. As a result, they often perform extensive revision work on the decision results [24]. Therefore, a considerable amount of structured historical data accumulated by the SCADA system for power data acquisition and monitoring control is the product of a combination of the dispatch automation system and dispatcher experience. Simultaneously, with increase in the uncertainty factors in the new power system, long-term operating scenarios exhibit massive high-dimensional and nonlinear correlation characteristics. However, traditional numerical statistical analysis based on principal components is limited by its ability to handle only a single data form and cannot conduct global analysis of all participants in the entire regional power system. Consequently, the deep extraction of information contained in power grid operating scenarios is hindered.

The advancement of computer application technology and the increasing popularity of online social media have made images an essential form of data owing to their ability to convey information with rich expressiveness and intuitiveness. Several engineering applications based on image data have accelerated the continuous maturity of computer vision technology and have assisted in solving various cross-domain bottleneck problems. Therefore, in this study, a GASF-based power system operating scenario image encoder was designed, which converts the power system operating scenarios into spatial image data. This encoder can generate a power grid operating scenario image dataset while preserving the experience of the dispatchers.

**Figure 1** shows the process of converting an operating scenario sequence into a scenario image, which includes three main steps. First, different operating scenario sequences are were represented by polar coordinates. The corresponding Gramian matrix was then generated and encoded to generate power system operating scenario images.

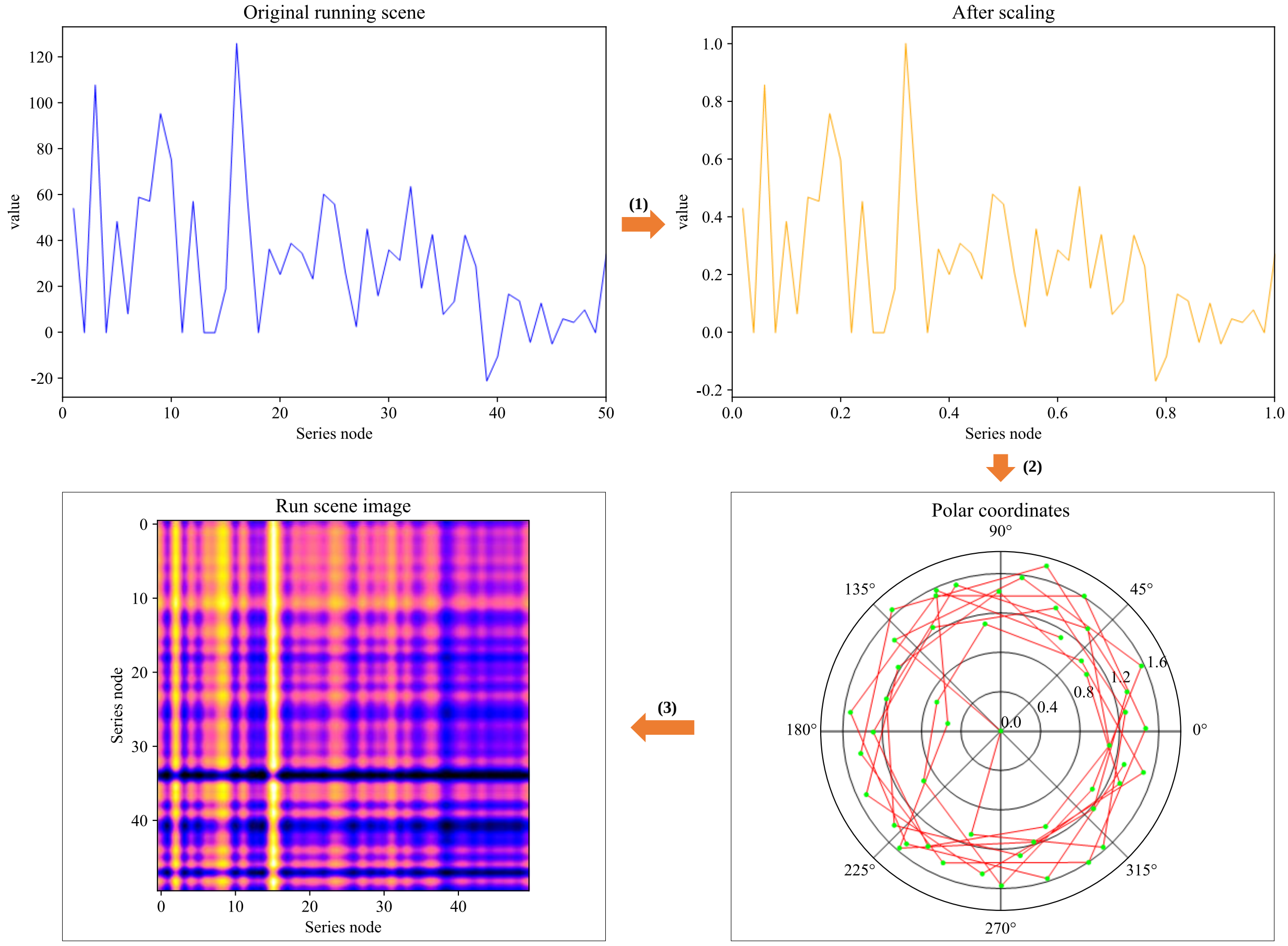

Fig. 1. Principle of operating scenario image encoder

(1) Normalization of the operating scenario sequence

According to Equation (1), each data point in the operating scenario is not a vector but a scalar; therefore, it cannot be directly used to construct a Gramian matrix. Simultaneously, considering that the discharge state of load-side energy storage nodes under the same operating scenario is marked as a negative value, the sequence data $s_i = \{G_{i,1}, \cdots, G_{i,g}, R_{i,1}, \cdots R_{i,r}, L_{i,1} \cdots, L_{i,l}\}$ of the time snapshot $i$ is normalized to the interval $[-1,1]$ using Equation (3):

$$\tilde{s}_i = \frac{s_i}{|\max|} \tag{3}$$

where $|\max|$ is the absolute value of the maximum value in the entire sequence.

(2) Polar coordinate transformations

The normalized values are converted into polar coordinates, and spatial dimension information is added, so that the running snapshot sequence index point is converted into the correlation between the node and other nodes. The equation used is as follows:

$$\begin{cases} \theta_j = \arccos(\tilde{s}_i^j), -1 \le \tilde{s}_i^j \le 1 \\ r_j = \dfrac{j}{m}, j \in [1, m] \end{cases} \tag{4}$$

The encoding process involves a polar coordinate transformation, which enables both aspects of the information to be included without any loss. From a mathematical perspective, it can be regarded as a bijective function with a one-to-one correspondence before and after transformation.

(3) Gramian matrix

According to Equation (4), GASF defines a special inner product calculation formula as follows:

$$\langle \tilde{s}_i^1, \tilde{s}_i^2 \rangle = \cos(\theta_1 + \theta_2) \tag{5}$$

Based on Equation (6), the "inner product" between the data points in the operating scenario sequence is the cosine of the sum of the polar angles after converting them into polar coordinates. The Gramian matrix for the entire operating scenario is expressed as Equation (6).

$$GM = \begin{bmatrix} \cos(\theta_1 + \theta_1) & \cos(\theta_1 + \theta_2) & \cdots & \cos(\theta_1 + \theta_m) \\ \cos(\theta_2 + \theta_1) & \cos(\theta_2 + \theta_2) & \cdots & \cos(\theta_2 + \theta_m) \\ \cdots & \cdots & \cdots & \cdots \\ \cos(\theta_m + \theta_1) & \cos(\theta_m + \theta_2) & \cdots & \cos(\theta_m + \theta_m) \end{bmatrix} \tag{6}$$

This equation represents the matrix product of $m \times m$. In the running scenario, each node was encoded as a matrix with the geometric dimensions of $1, 2, \cdots, m$. Further expansion of the inner product measure in Equation (6) yields:

$$\begin{aligned} &\cos(\theta_1 + \theta_2) = \cos(\arccos(\tilde{s}_i^1) + \arccos(\tilde{s}_i^2)) \\ &= \cos(\arccos(\tilde{s}_i^1)) \cdot \cos(\arccos(\tilde{s}_i^2)) - \sin(\arccos(\tilde{s}_i^1)) \cdot \sin(\arccos(\tilde{s}_i^2)) \\ &= \tilde{s}_i^1 \cdot \tilde{s}_i^1 - \sqrt{1 - \left(\tilde{s}_i^1\right)^2} \cdot \sqrt{1 - \left(\tilde{s}_i^2\right)^2} \end{aligned} \tag{7}$$

Through a high-dimensional space transformation, $s_i$ contains amplified information, rendering it easier to extract more refined hidden operating information. Moreover, the data format of the operating scenario image provides a more feasible method for information extraction.

## 4 DTSAs for typical operating scenario extraction

Most existing methods for extracting information from power system operational snapshots follow the mathematical paradigm of feature selection and traditional clustering algorithms, with no interaction between them. This destroys the complete historical dispatching experience of dispatchers and renders it impossible to perform cluster analysis on the system operational snapshot over long time scales. Based on the mathematical representation of typical operating scenarios in Section 2, we propose DTSAs to extract the typical operation scenarios of new power systems. DTSAs comprises two main steps: operating scenario image feature extraction and typical operating scenario generation, with operating scenario images as the input data, as described in Section 3. The framework of the proposed scheme is shown in **Fig. 2**.

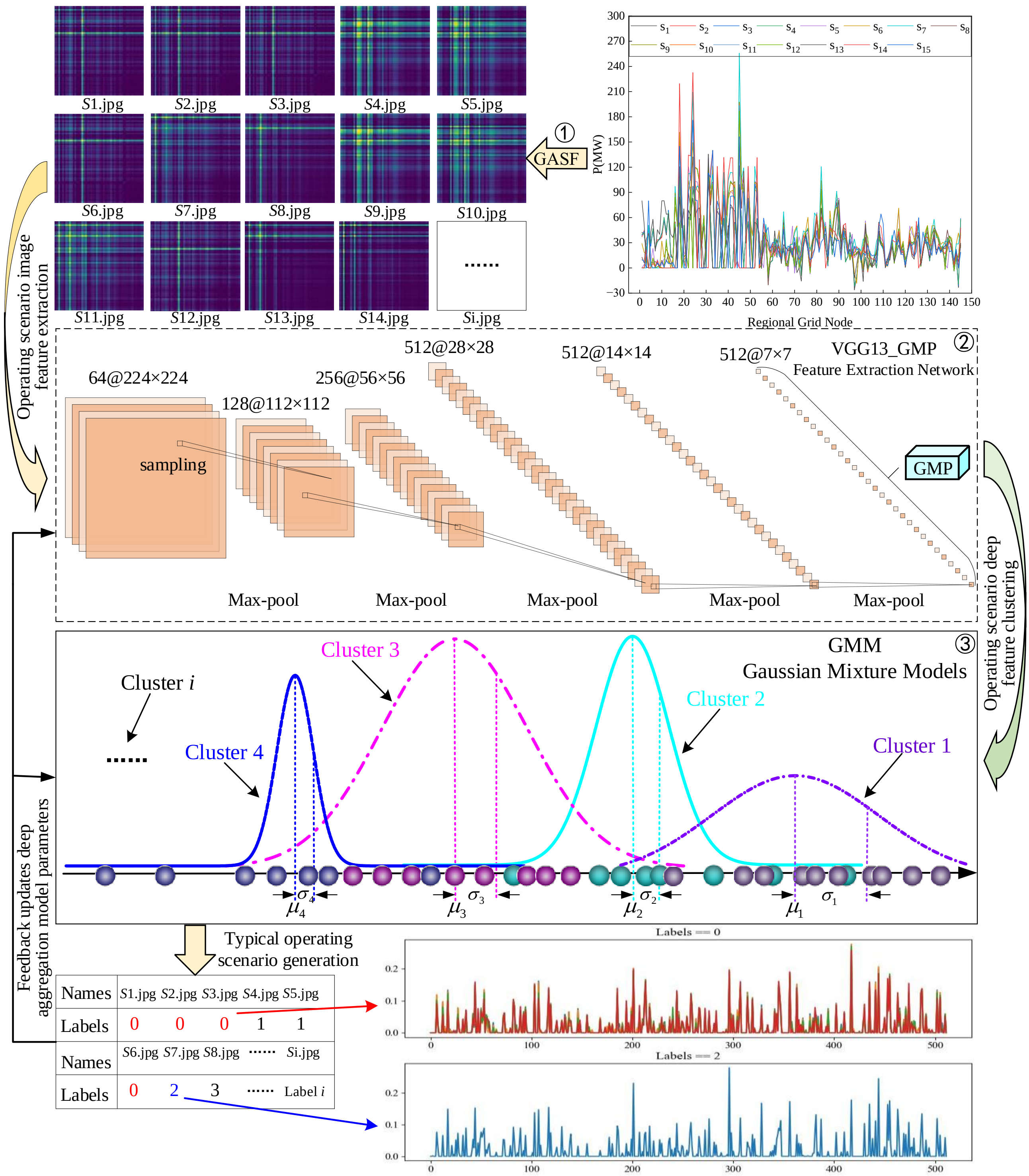

Fig. 2. Overall framework of DTSAs. (The resulting scenario aggregation during the model training process is responsible for updating the weight parameters of the convolutional layers and aggregation algorithms in steps 2 and 3, respectively, and the updated model further generates typical operating scenarios. The two form an iteration until the aggregation result reaches the optimal performance indicator.)

### *4.1 Basic concepts of deep time series aggregation*

Clustering is among the most important research branches in the field of data mining. It involves differentiating original data into different subsets based on similarity measurements. Deep aggregation introduces deep learning into the clustering field and uses clustering results to guide neural networks in reverse to learn more prominent features of clustering objects. This effectively overcomes the bottleneck of shallow (traditional) clustering methods, which cannot handle high-dimensional data types [25]. Its general paradigm can be expressed as

$$Loss = \lambda_1 L_{net} + \lambda_2 L_{clustering} \tag{8}$$

where $Loss$ is the combination of neural network loss $L_{net}$ and clustering loss $L_{clustering}$, and $\lambda_1 \geq 0$ and $\lambda_2 > 0$, represent the weight coefficients of the deep clustering model, with $\lambda_1 + \lambda_2 = 1$. When $\lambda_1 = 0$, the model is equivalent to a traditional clustering model.

In the case of massive high-dimensional data, there is often a conflict between feature dimensionality reduction and information integrity. The concept of deep time series aggregation effectively solves this problem by forming a closed loop between learning high-quality features and improving the performance of the clustering algorithms. **Fig. 3** compares the principles of deep and traditional clustering methods for high-dimensional data curves.

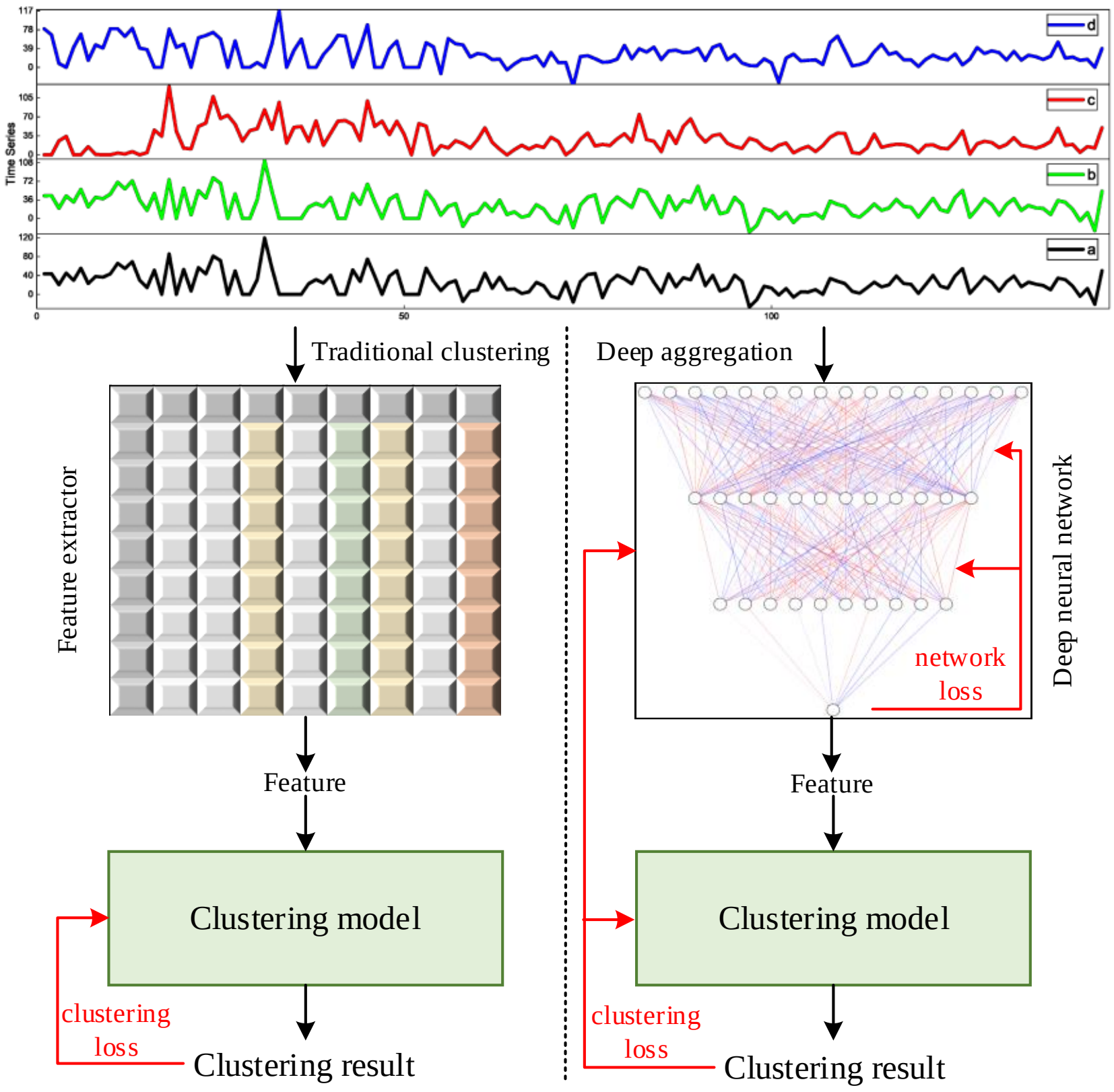

Fig. 3. Schematic comparison of deep aggregation and traditional clustering methods

### *4.2 Operating scenario image feature extraction network*

Image recognition is a crucial aspect of computer vision applications as it involves analyzing image categories by extracting feature information from images. The effectiveness of feature extraction directly affects the effectiveness of image recognition. Convolutional neural networks (CNNs) are a powerful deep learning method that consists of convolutional and pooling layers, and they have been successful in solving many complex pattern recognition problems [26]. The operating scenario transformation operation in Section 3 effectively realizes the image-form expression of power system time-series operating data, thereby providing a new path for the analysis and extraction of typical operating scenarios in the system while enriching the content of data representation. Therefore, this study innovatively designed a power system operating scenario image feature extraction network based on VGG13_GMP, as shown in **Fig. 4**. The network comprises an input layer, 13 convolutional layers, 5 pooling layers, and a feature output layer. Max-pooling was used between layers, and the ReLU function was used for all hidden layer activation units. In addition, global maximum pooling was employed for the feature output layer.

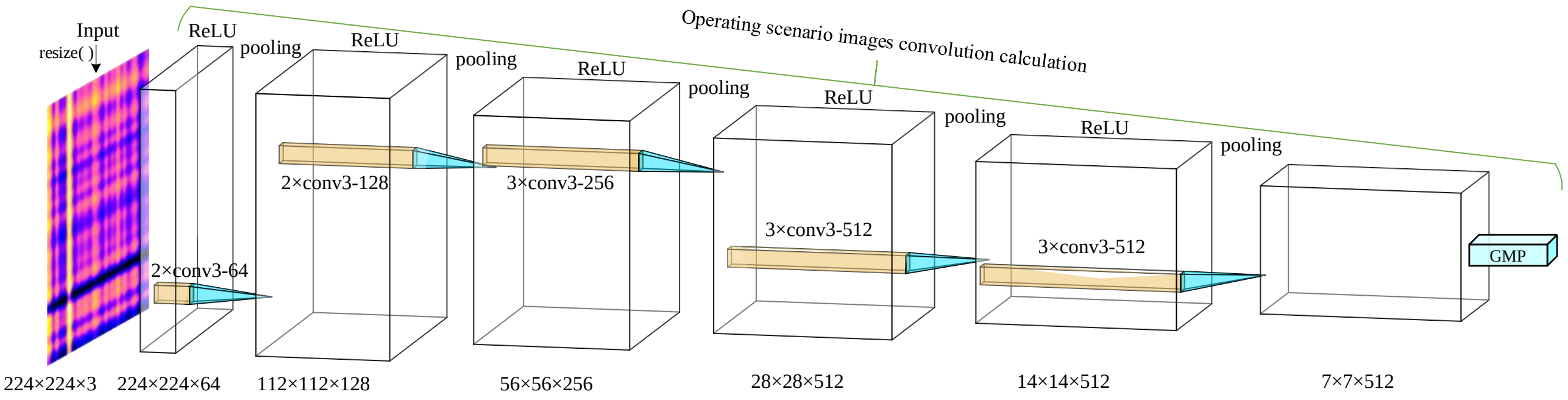

Fig. 4. Structure of power system operating scenario image feature extraction network based on VGG13_GMP

The specific process of operating scenario image feature extraction is as follows:

First, the network input layer image size was set to $224\times224\times3$, thus, the resize() function was initially used to adjust the size of the operating scenario image. Following adjustment, the image was convolved twice with 64 channels and $3\times3$ convolutional kernels of size 3. The calculation formula is expressed as Equation (9):

$$n_{out}=\frac{n_{in}+2pd-k_{size}}{sd_{conv}}+1 \tag{9}$$

Subsequently, a maximum pooling layer was added after each group of convolutional layers to downsample the input image to the convolutional layer. The calculation formula is shown in Equation (10). The pooling operation compressed the input feature, thus simplifying the complexity of the convolutional network calculation. Moreover, it maintained the rotational, translational, and scaling invariance of features.

$$p_{out}=\frac{p_{in}-f_{size}}{sd_{pool}}+1 \tag{10}$$

The entire power system operating scenario image feature extraction stage involved 13 convolutions and 5 pooling layers, as shown in **Fig. 4**. The changes in the feature map size before and after the calculation were obtained using Equations (9) and (10), respectively. Meanwhile, the feature extraction model used multiple small convolutional kernels of size $3\times3$ to increase the network depth through multiple nonlinear layers. This enabled the learning of extreme operating modes under complex power system operating conditions.

Finally, although the designed power system operating scenario image feature extraction network was based on the basic VGG16 [27] model, the fully connected and classification layers were discarded, and a 2D global maximum pooling operation, as shown in **Fig. 5**, was used instead. Thus, to avoid the issue of excessive weight parameters in the first fully connected layer, the most prominent features of the entire feature extractor were retained and utilized as inputs for the aggregation algorithm, which is described in the next Section. Compared with VGG16, the GMP at the end of the model does not contain any parameters, the model size is reduced by 473 MB, and the number of parameters is reduced by 123 633 664. This effectively reduces the computational complexity of the feature extraction stage.

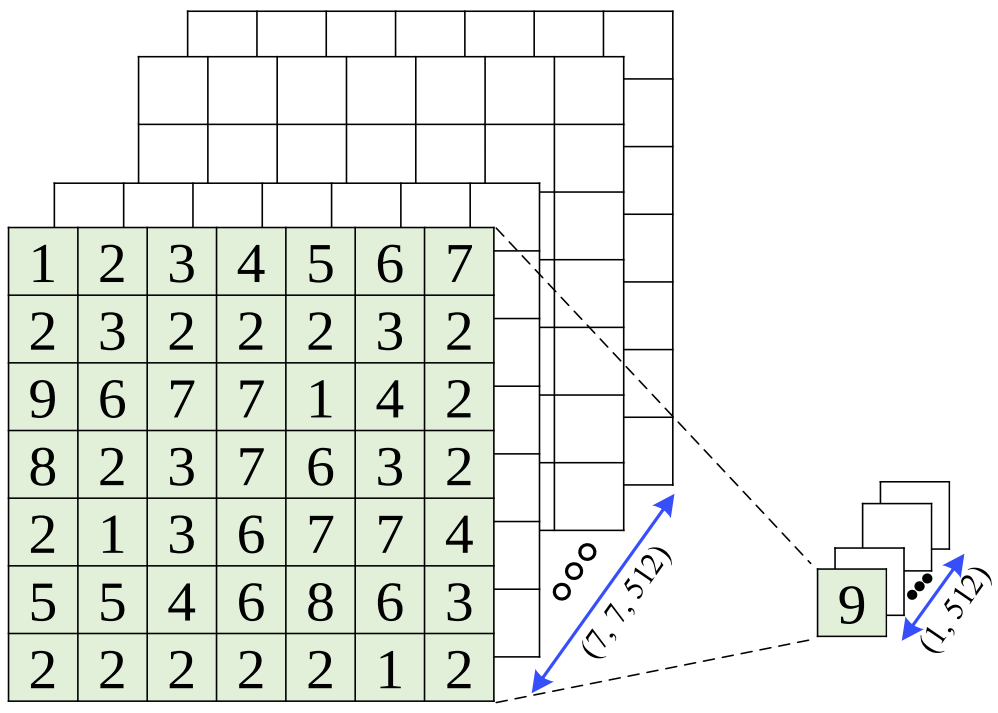

Fig. 5. Output layer of the feature extractor

***4.3 Typical operating scenario generation algorithm***

Traditional distance-based hard clustering algorithms are ineffective because the generated feature map sequences of power system operating scenarios exhibit typical distribution characteristics. Gaussian Mixture Models (GMM) are generative models that aim to discover a mixed representation of multidimensional gaussian model probability distributions capable of fitting data distributions of any shape of operating scenario [28]. This section introduces an iterative feature aggregation algorithm for generating typical operating scenarios. The algorithm's underlying concept is that utilizing fine-grained features enhances the GMM's ability to produce accurate clustering results. These clustering results, in turn, guide the VGG13_GMP to update the network weight coefficients, facilitating the learning of better features. The seamless connection between the two processes significantly improves the clustering performance of the power system operating scenarios.

The algorithm assumes that the operational snapshot contained in each typical operating scenario follows a separate multivariate gaussian distribution and that the entire operational snapshot dataset follows a mixture of Gaussian distributions. Specifically, the distribution of operating scenarios can be represented by Equation (11).

$$p(s_i) = \sum_{k=1}^{K} w_k N(s_i \mid \mu_k, \Sigma_k) \tag{11}$$

where $w_k$ represents the probability that an operational snapshot belongs to the $k$ th class, and the sum of the total weights equals 1. Further, $N(\bullet \mid \mu, \Sigma)$ is a Gaussian distribution with mean $\mu$ and covariance matrix $\Sigma$ . By maximizing the likelihood function of $p(s_i)$ , $w, \mu, \Sigma$ can be solved iteratively as

$$L_{clustering}(w, \mu, \Sigma) = -\ln p(S \mid w, \mu, \Sigma) = -\sum_{i=1}^{t} \ln \left\{ \sum_{k=1}^{K} w_k N(s_i \mid \mu_k, \Sigma_k) \right\} \tag{12}$$

In Equation (12), the negative log-likelihood function is added to convert it into a GMM clustering loss.

The VGG13_GMP deep neural network mapped the operating scenarios to the features. Based on Equation (12), the loss function of the deep clustering (NC) algorithm based on Gaussian mixture models is defined as follows:

$$\begin{cases} L_{NC}(\omega, w, \mu, \Sigma) = -\ln p(Z \mid w, \mu, \Sigma) \\ \qquad = -\sum_{i=1}^{t} \ln \left\{ \sum_{k=1}^{K} w_k N(z_i \mid w, \mu, \Sigma) \right\} \\ s.t.\ z_i = f_\omega(s_i) \end{cases} \tag{13}$$

From the perspective of probabilistic graphical models, $z_i$ can be considered as a new latent variable that is expected to infer the distribution of the original data more accurately through this latent variable. The optimization problem corresponding to this goal was solved using an expectation-maximization (EM) algorithm. The optimized parameters included $\theta = \{\omega, w, \mu, \Sigma\}$ , which $\omega$ is the mapping model parameter of the latent variable $z_i$ , and $w, \mu, \Sigma$ are the GMM parameter.

Step 1: Initialize $\theta$ . Here, $\omega$ uses pretrained imageNet network convolutional layer weights and the other parameters use the results obtained from running k-means. Consequently set the number of components for the Gaussian mixture model.

Step 2: Perform the E-step. Fixing the existing parameters $\theta$ , calculate the probability of each running scenario $s_i$ belonging to each cluster using Equation (14), which is the posterior probability. The higher the value, the more likely the running scenario is assigned to the correct cluster, and vice–versa.

$$p(z_{ik}) = \frac{w_k N(z_i \mid \mu_k, \Sigma_k)}{\sum_{j=1}^{K} w_j N(z_i \mid \mu_j, \Sigma_j)} \tag{14}$$

Step 3: After completing the E-step, calculate the parameters $\theta$ that maximize the probability of the running scenario, that is, perform the M-step. Use the weighted probabilities of data points to calculate the new parameters.

$$\begin{cases} \omega^* = \arg\max_{\omega} F(p^{(k)}(z_{ik}), \omega) \\ w_k^* = \dfrac{\sum_{i=1}^{t} p(z_{ik})}{t} \\ \mu_k^* = \dfrac{\sum_{i=1}^{t} p(z_{ik}) z_i}{\sum_{i=1}^{t} p(z_{ik})} \\ \Sigma_k^* = \dfrac{\sum_{i=1}^{t} p(z_{ik})(z_i - \mu_k)(z_i - \mu_k)^T}{\sum_{i=1}^{t} p(z_{ik})} \end{cases} \tag{15}$$

Step 4: Repeat iterations 2 and 3 until convergence is achieved.

### *4.4 Computational complexity of DTSAs*

DTSAs are an end-to-end deep clustering method. Its computational complexity is divided into two aspects:

1) Computational complexity of feature extraction stage: The computational complexity of this stage depends on the architecture, depth, and number of parameters of the neural network. Feature extraction of running snapshot images is implemented by 13 convolutional layers of VGG13. The computational complexity consists of the size of the input image $H_{in} \times W_{in}$, the size of the convolution kernel $k_{size}$, the number of convolution kernels $n_{\text{kernel}}$, and the size of the output feature map $H_{out} \times W_{out}$. Specifically, it can be expressed as Equation (16):

$$O_{conv} = H_{out} \times W_{out} \times n_{\text{kernel}} \times k_{size} \times k_{size} \tag{16}$$

where $H_{out}$ and $W_{out}$ are calculated by Equation (9).

2) Computational complexity of the clustering stage: The computational complexity of this stage depends on the clustering algorithm used. In this paper, the computational complexity of GMM is affected by the number of operating snapshots $n_{sample}$, the number of Gaussian components $K$, the dimension of each Gaussian component $D$ and the number of iterations of the EM algorithm $N_{epoch}$, which can be expressed as Equation (17):

$$\begin{cases} O_E = n_{sample} \times K \times D \\ O_M = n_{sample} \times D + n_{sample} \times D^2 + n_{sample} \times K \\ O_{clustering} = (O_E + O_M) \times N_{epoch} \end{cases} \tag{17}$$

where $O_E$ is the computational complexity of E-step. $O_M$ is the computational complexity of the M-step, which is divided into the computational complexity of updating the mean, the computational complexity of updating the covariance matrix, and the computational complexity of updating the mixing coefficient. $O_{clustering}$ is the total computational complexity of the entire clustering stage.

Therefore, the computational complexity of DTSAs is $O_{conv} + O_{clustering}$.

## 5 Experiments

This section presents experimental analyses based on a real power grid at a provincial level in China. The effectiveness and superiority of our method were examined based on the qualitative and quantitative evaluation results. In addition, we designed experiments to examine the robustness of the DTSAs for three grid scenarios with different new energy penetration rates. The case study was conducted using Python 3.7 on a standard PC equipped with an Intel Xeon Gold 5118 CPU with a clock frequency of 2.30 GHz, 128 GB RAM, and an NVIDIA Quadro P5000 independent graphics card.

### *5.1 Description of experimental data and case scenarios*

The experimental data were a time-series simulation dataset generated by considering various control methods such as conventional units, new energy units, adjustable loads, and energy storage for the province in 2019. **Table 1** presents complete information on the case system. Among them, the maximum output of new energy throughout the year exceeded 1323 MW, accounting for 59.97% of the total power generation output at that time. Further, the proportion of adjustable load was 10%, and that of energy storage was 5%.

Table 1. Complete information description of the case system.

| | | Area 1. | | Area 2. | | Area 3. | |
|---|---|---|---|---|---|---|---|
| Number of nodes. | | 42 | | 45 | | 39 | |
| Generation nodes. | Conventional thermal power generation | 19 | 18 | 20 | 16 | 15 | 2 |
| | New energy generation. | | 1 | | 4 | | 13 |
| Load nodes. | Conventional load. | 34 | 33 | 30 | 27 | 27 | 19 |
| | Including energy storage participation. | | 1 | | 3 | | 8 |
| Number of lines. | | 57 | | 59 | | 50 | |
| Peak load (MW) | | 1193 | | 1400 | | 1094 | |
| Total installed capacity (MW) | Thermal power installed capacity | 2670 | 2570 | 2863 | 2483 | 1390 | 275 |
| | Installed capacity of new energy | | 100 | | 380 | | 1115 |

Based on the grid structure characteristics of the power system in this province, this study designed three scenarios of new energy penetration rates: low, medium, and high access ratios. Consequently, the form of the power grid in each stage of the development of the new power system was deduced. **Fig.6** shows the network topology under different new energy penetration rate scenarios. The red dots represent the thermal power units, green dots represent the new energy units, blue buses indicate that energy storage is included in the load, black buses carry ordinary loads (including adjustable loads), yellow lines represent the regional boundaries, and external lines represent the interconnection lines with other regions.

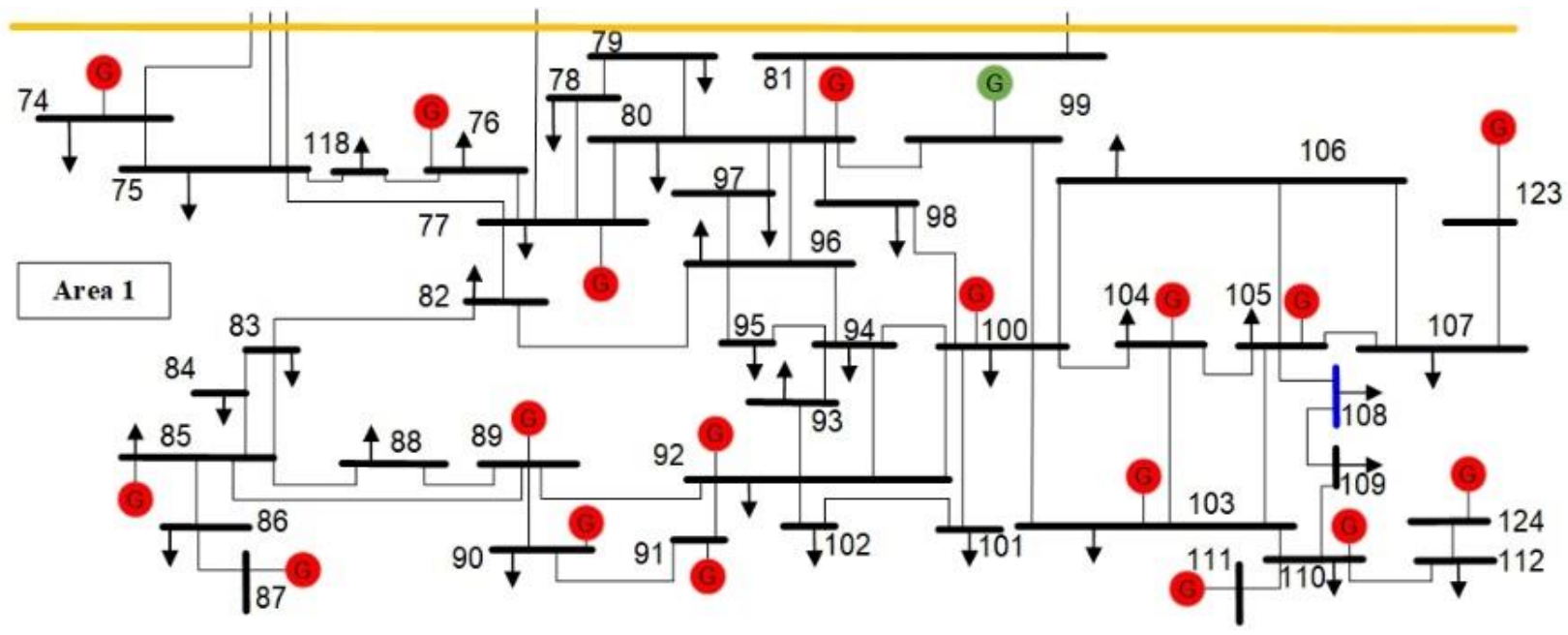

(a) Low access ratio scenario

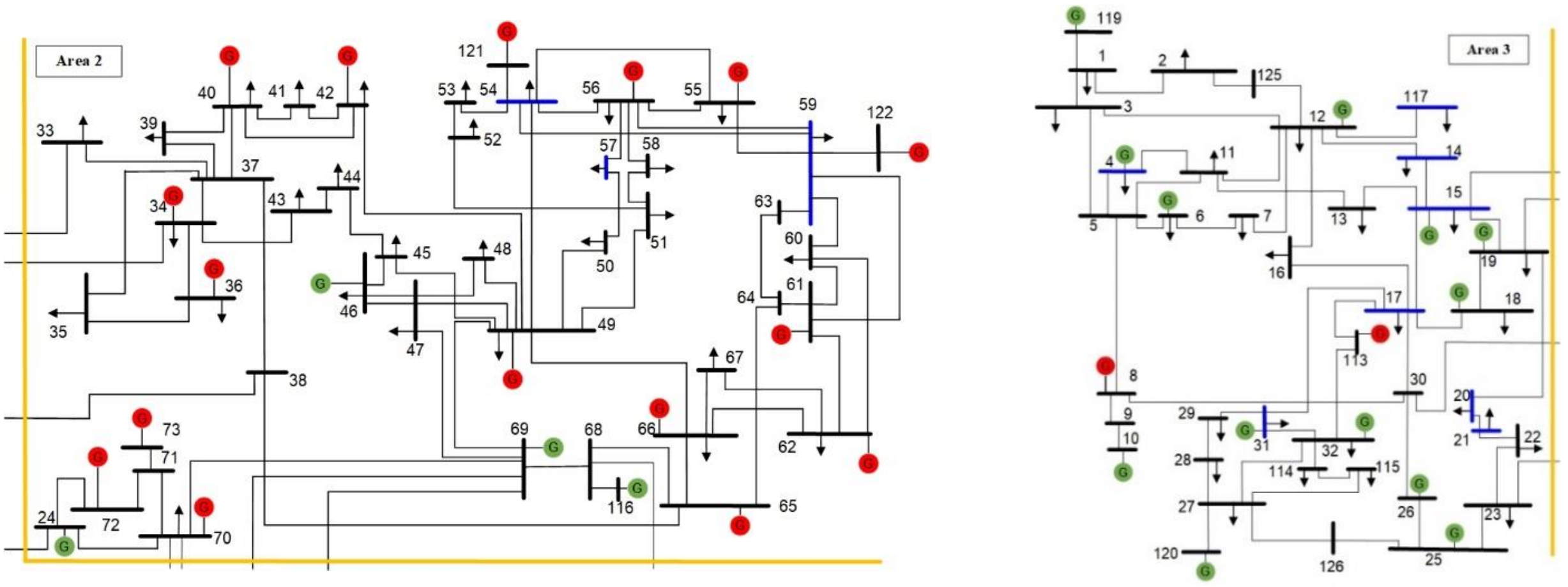

(b) Medium access ratio scenario

(c) High access ratio scenario

Fig. 6. Grids under different new energy penetration rate scenarios

**Figure 6** clearly shows that energy storage plays an important role in the grid operation during the construction of a new power system. In the low-access-ratio scenario, which represents the traditional power system operation mode, the generator units themselves have strong adjustability and adaptability, and the demand for energy storage is small and primarily serves as emergency frequency regulation. As shown in **Fig. 6(c)**, under a high proportion of new energy access scenarios, the grid operation was gradually dominated by wind and solar power. In addition, the temporal fluctuation increased significantly, requiring more energy storage to participate in the regulation of the operation mode. This further illustrates the significance of the proposed method for extracting refined operation scenarios in the context of the new power system.

After considering the boundary conditions outlined in the above case scenario, historical operational snapshot data from this province's power system with new energy integration was utilized with a time resolution of 5 minutes to verify the feasibility and effectiveness of DTSAs in various stages of the new power system's evolution. The following presents the specific case analysis and conclusions.

### *5.2 Experimental results and analysis*

#### 5.2.1 Scenarios with low proportion of new energy access

**Fig. 7** illustrates the extraction of six typical operating scenarios from the power system operational snapshots presented in **Fig. 6(a)**. Each scenario represents a typical operating state that includes the optimal matching status of all the commissioned equipment in the regional system. These scenarios satisfy real-time power balance and equipment safety constraints while comprehensively considering multiple objectives, such as new energy consumption, power system operating costs (unit operating costs, adjustable resource cost), and penalties for power flow limit violations.

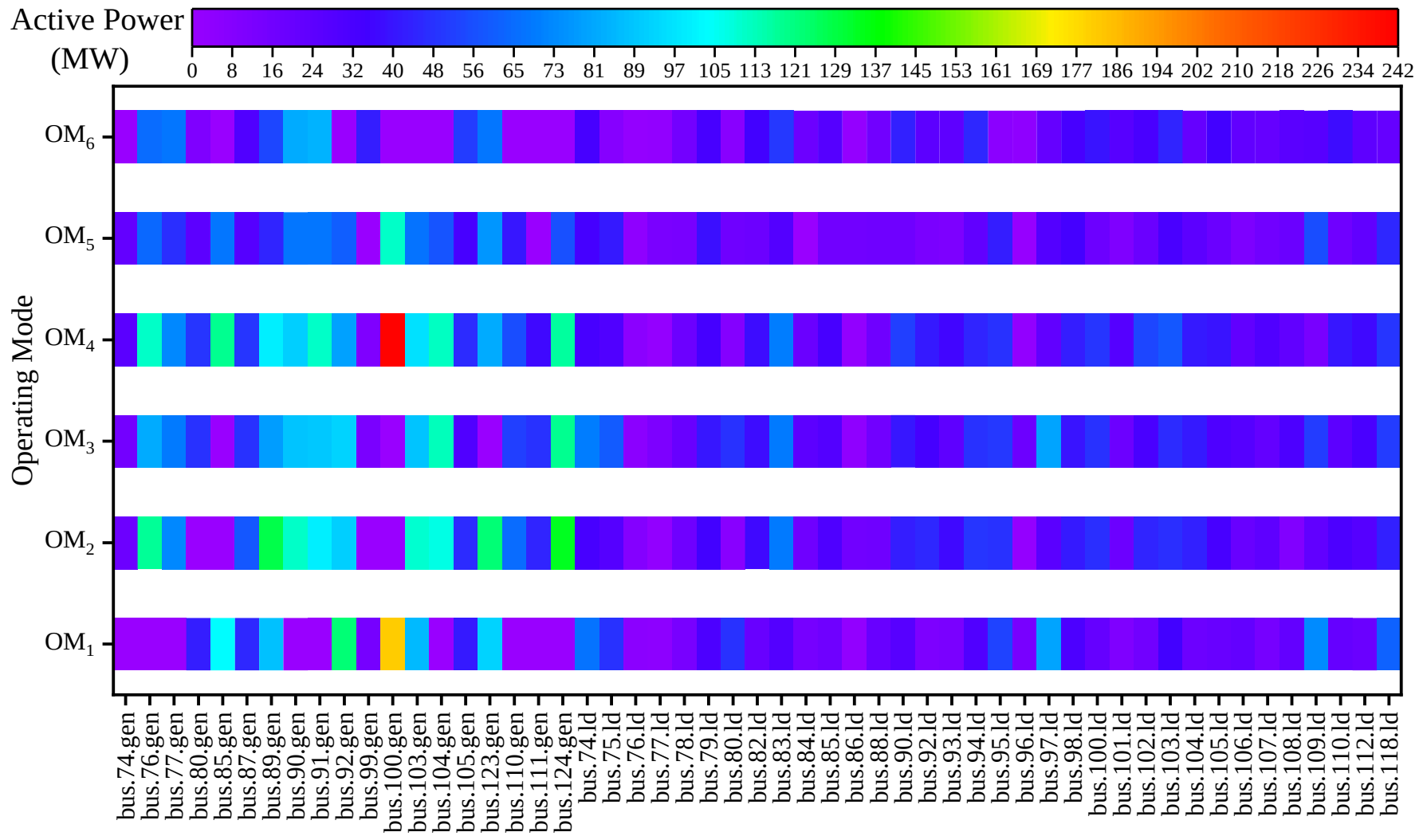

Fig. 7. Typical operation scenarios under low proportion of new energy integration

The low integration scenario shown in **Fig. 7** indicates that the differences between load nodes in area $OM_i$ are not significant. The primary difference lies in the coordinated scheduling of the power generation nodes. This can be attributed to the traditional power system operating in a "power follows load" mode, where loads are considered as rigid and do not participate in the scheduling mode adjustment. The power generation side, which is primarily composed of thermal power, is relatively flexible in control and produces different operation plans through coordination, which is affected by factors such as load demand, unit maintenance, transmission line load constraints, and economic constraints.

**Fig. 8** shows the new energy output of each typical operation scenario shown in **Fig. 6(a)**. Notably, there is no typical operation scenario wherein the rated capacity output of new energy appears. This indicates that the absorption rate of a few new energy nodes is not an important scheduling constraint indicator in this scenario.

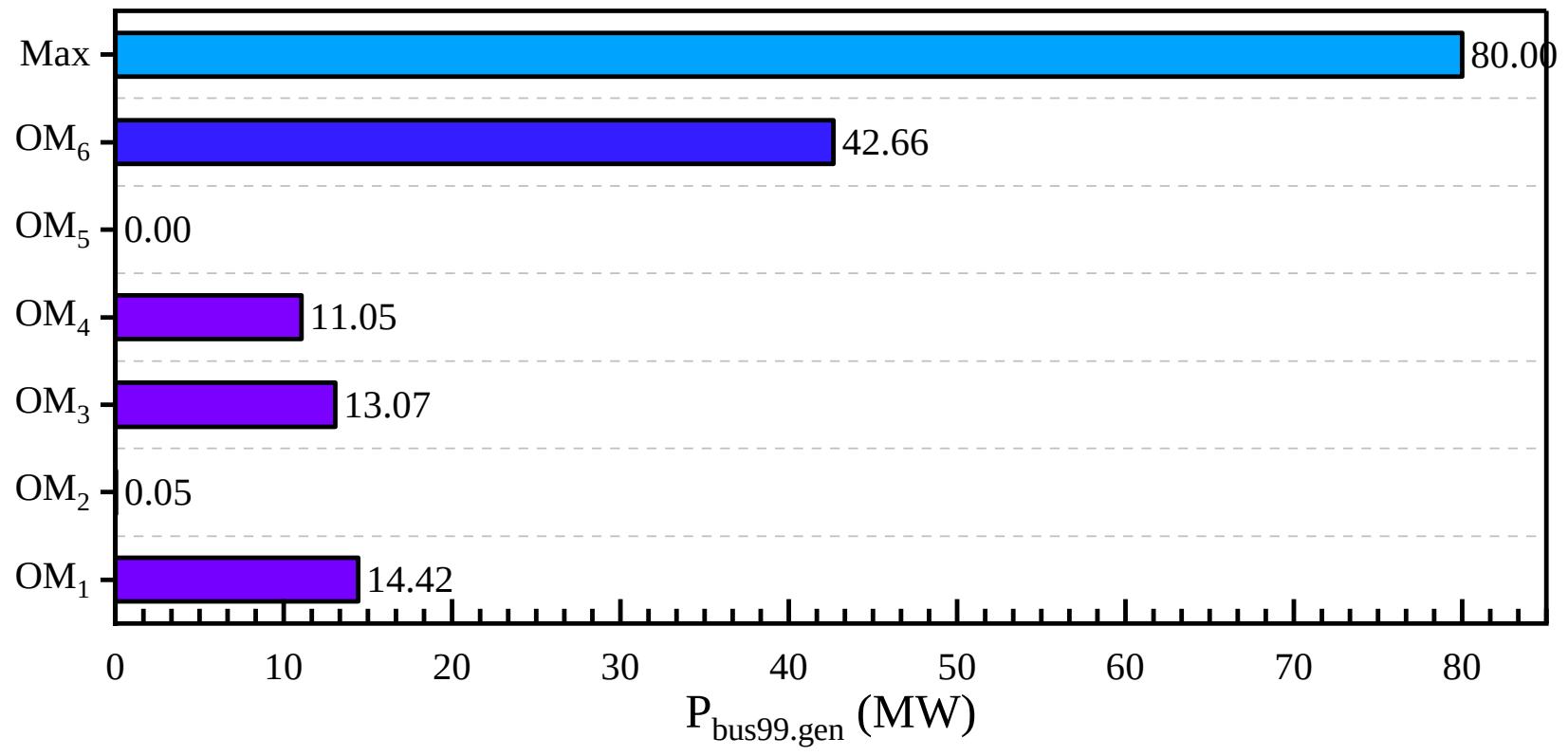

Fig. 8. New energy node output of each typical operation scenario under low integration scenario

### 5.2.2 Scenarios with medium proportion new energy access

The regional power grid in **Fig. 6(b)** reflects the initial stage of the construction of a new power system, where new energy nodes account for 20% and the installed capacity accounts for approximately 13.27%. Upon analyzing the typical operation scenarios extracted from **Fig. 9**, it was observed that there was a gradual shift in focus towards the consumption of new energy as the energy dispatching process progressed. Moreover, typical operation scenarios that consider the operational characteristics of new energy gradually appeared.

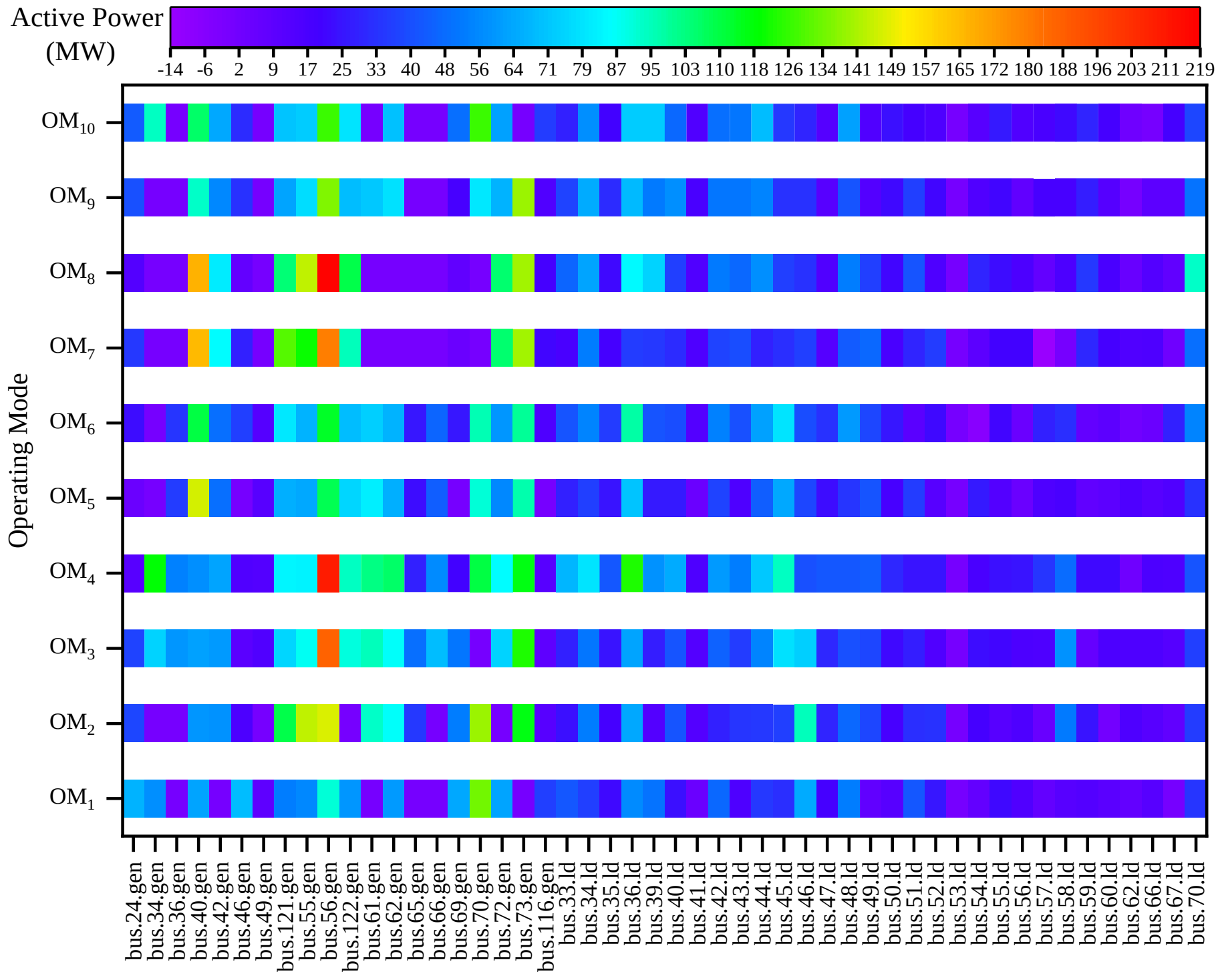

Fig. 9. Typical operation scenarios under medium proportion of new energy integration.

As shown in **Fig.10**, the new energy outputs corresponding to the 10 operation scenarios under the medium integration scenario are provided. Among them, $OM_1$ and $OM_5$ correspond to typical operation scenarios of a regional power system under full and non-dispatching conditions of new energy nodes, respectively.

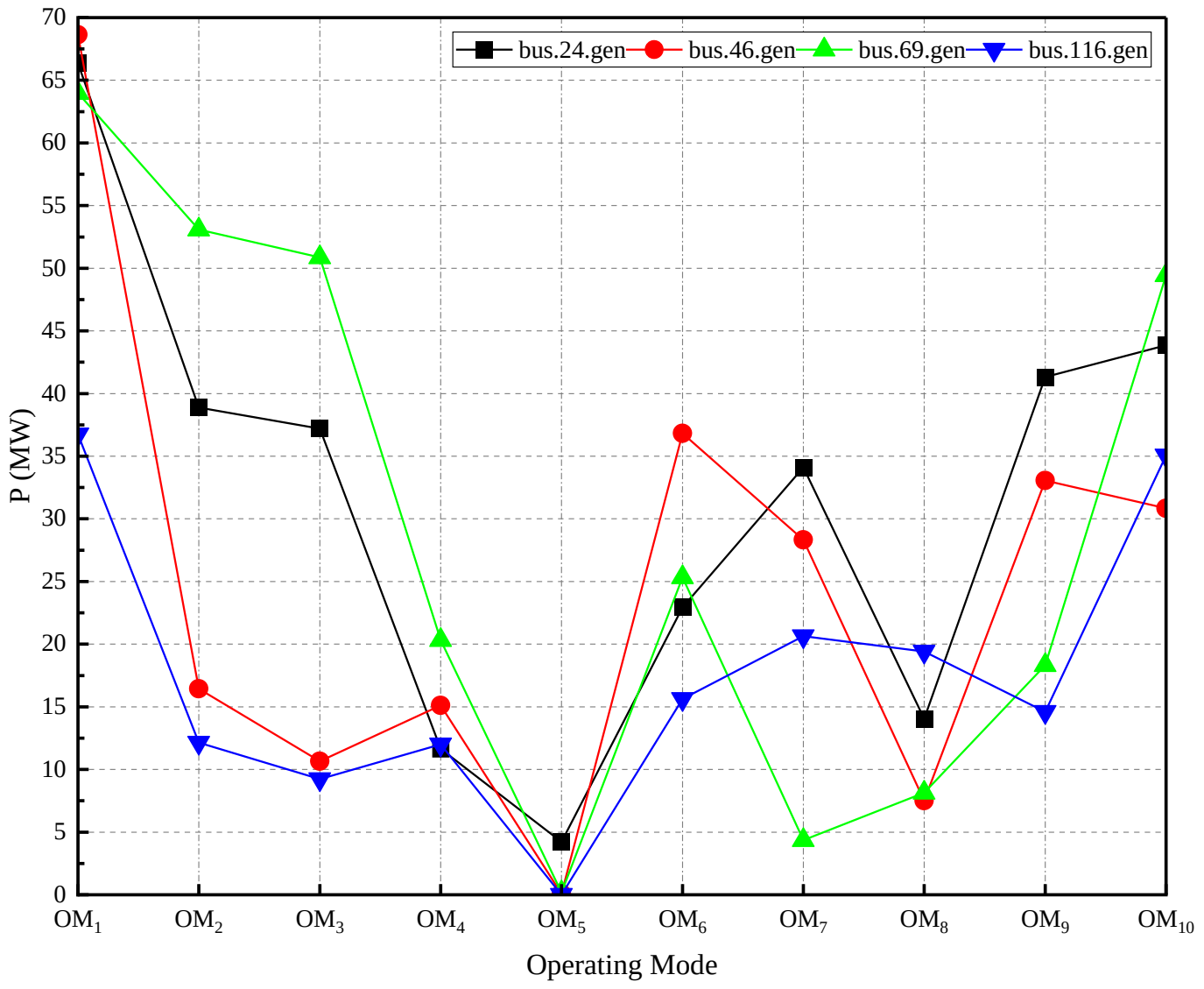

Fig. 10. New energy node output of each typical operation scenario under medium integration scenario.

**Figs. 11(a)** and **11(b)** show the continuous operation of all power generation nodes under Area 2 for two typical operation scenarios: $OM_1$ and $OM_5$. At time $OM_1$, new energy nodes in the area exist in a full dispatch state close to the rated capacity, whereas the other thermal power nodes provide flexible adjustments to ensure safe and stable operation of the regional power system. Whereas, at time $OM_5$, owing to the natural conditions, the output of the new energy nodes is approximately zero, and the entire area is supplied with thermal power. An analysis of the power grid configuration in this scenario shows that traditional thermal power generation is still the dominant power source; however, it maximizes the consumption of new energy by adjusting the output of thermal power during the high output of new energy.

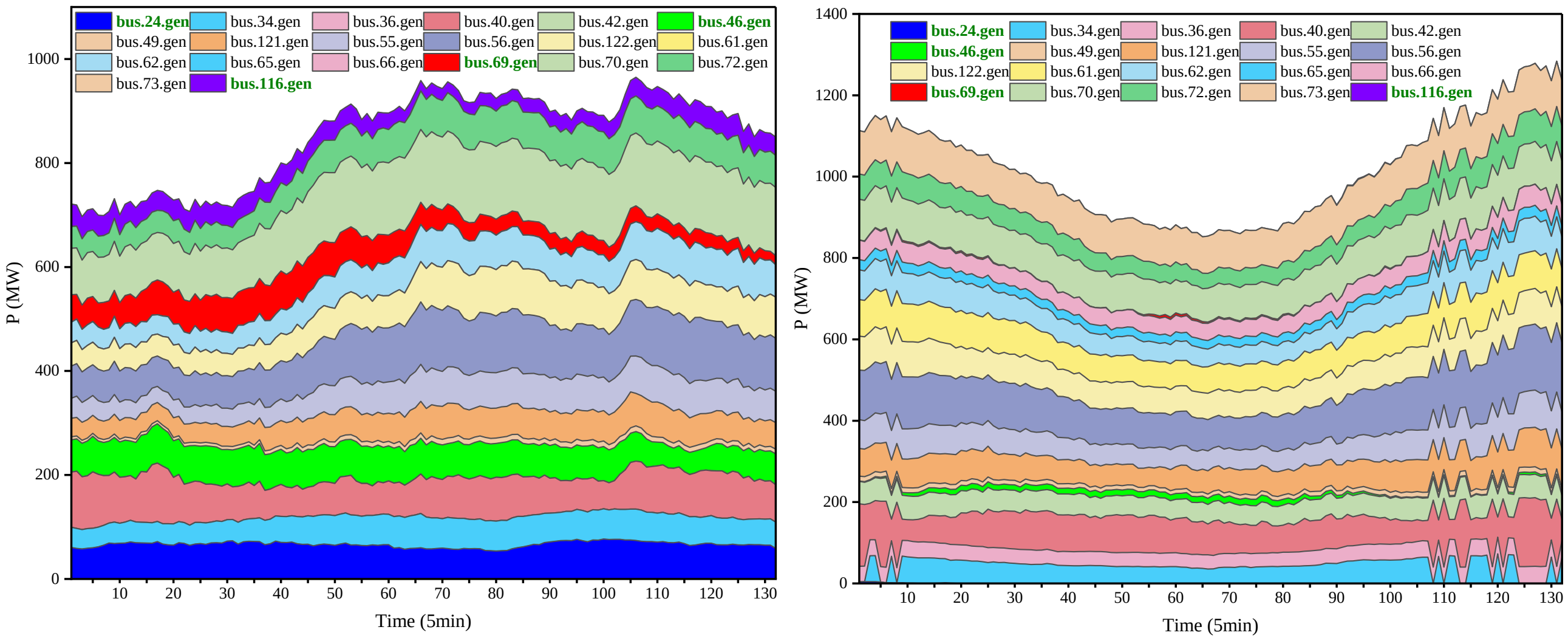

(a) New energy node full dispatch state

(b) New energy node non-dispatch state

Fig. 11. Typical operation scenarios considering new energy output under medium proportion of new energy integration.

### 5.2.3 Scenarios with high proportion of new energy access

**Fig.12** shows the aggregation results of the operation scenarios and the corresponding typical operation scenarios under a high proportion of new energy integration scenarios. Compared with **Figs. 7** and **9**, the number of typical operation scenarios significantly increased with an increase in the new energy integration ratio and energy storage nodes: from 6 in the low integration scenario to 19 in the high integration scenario. Specifically, four significant characteristics were observed: i) the clustering centers of various scenarios were considerably different based on the proposed method; ii) the high integration ratio of new energy led to a significant increase in the number of typical operation scenarios; iii) the distinction between typical operation scenarios included the involvement of energy storage and adjustable loads, such as $OM_4$ and $OM_{11}$; iv) regional interconnection lines played an important role in typical operation scenarios with large fluctuations in new energy, such as $OM_1$, $OM_8$, and $OM_{13}$.

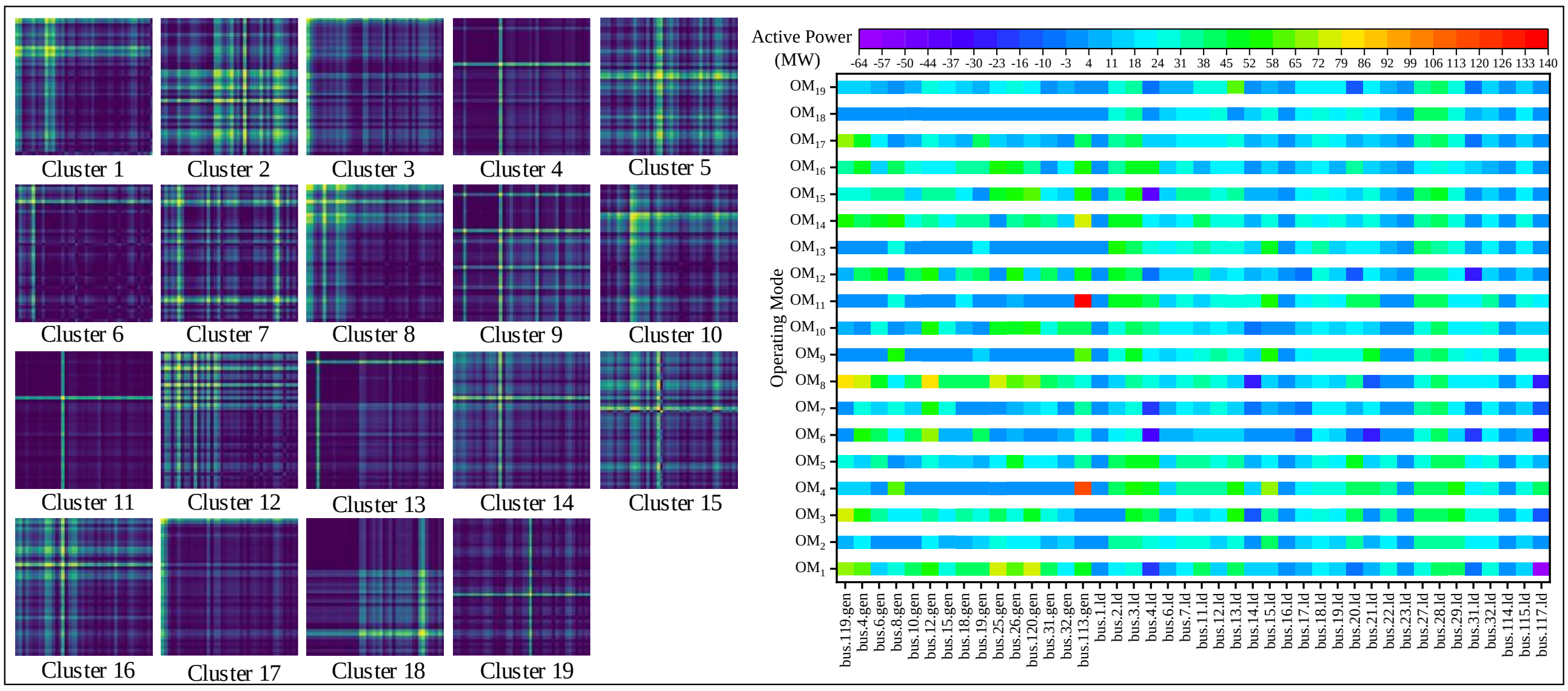

Fig. 12. Typical operation scenarios of the power system under high new energy penetration rate scenario

**Fig.13** analyzes the supply and demand characteristics of the 19 typical operation scenarios extracted. The energy storage node was placed on the $P_{generator}$ side in discharge mode and on the $P_{load}$ side in charging mode. The blue curve reflects the total power transmitted by all regional interconnection lines to maintain system balance.

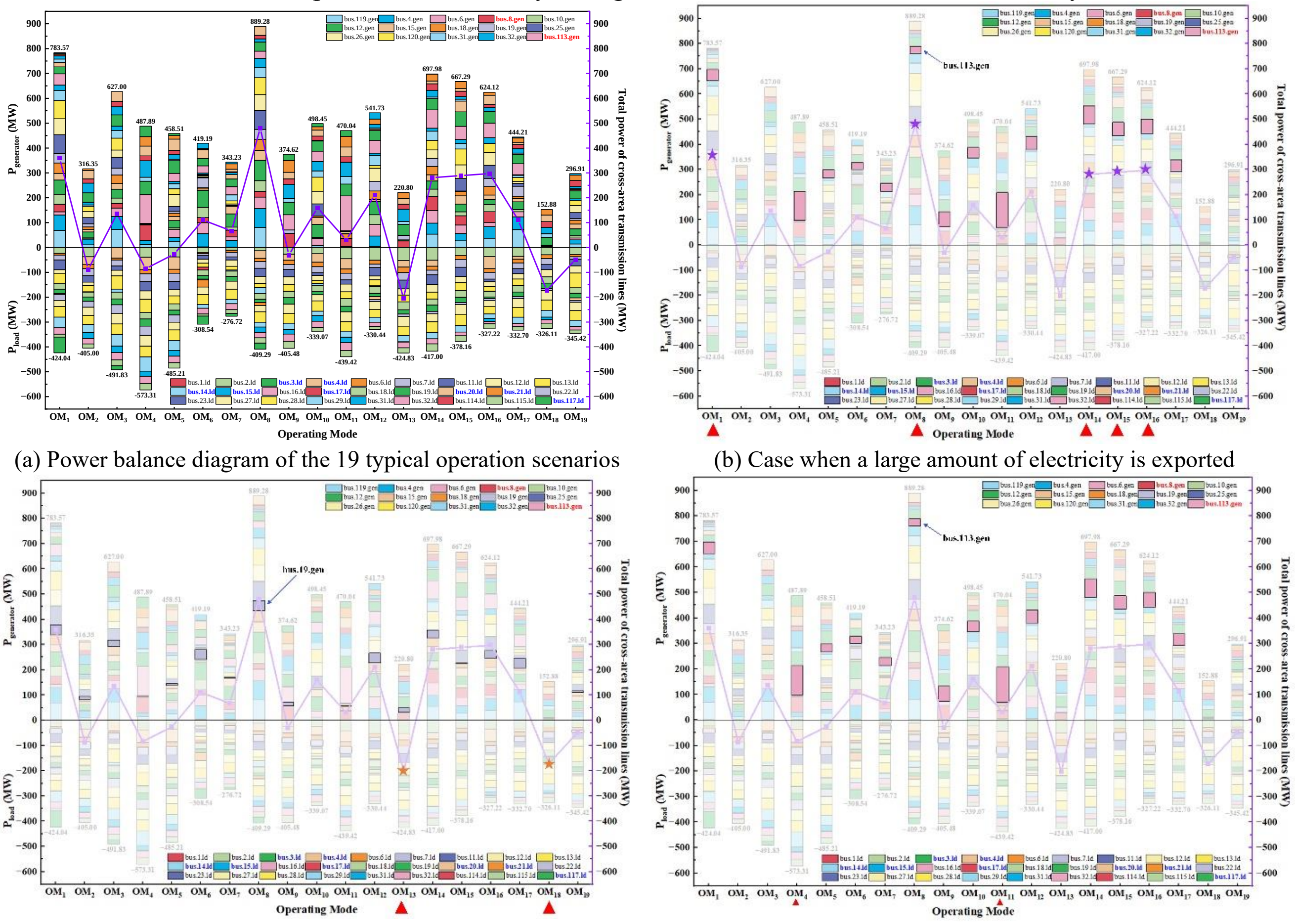

(a) Power balance diagram of the 19 typical operation scenarios
(b) Case when a large amount of electricity is exported
(c) Case when a large amount of electricity is absorbed
(d) Case when thermal power and energy storage jointly participate

Fig. 13. Power balance analysis of typical operation scenarios under high proportion of new energy integration scenario

Based on **Fig. 13**, the scenarios $OM_1$, $OM_8$, $OM_{14}$, $OM_{15}$, and $OM_{16}$ exhibit high levels of electricity export. Further, the analysis of Fig. 13(b) shows that above the 0 coordinate axis, bounded by bus_113_gen, the lower part is the new energy output in the region, and the upper part is the energy storage participation. Their common feature is that the new energy nodes have a high level of output. In contrast, scenarios $OM_{13}$ and $OM_{18}$ have low or almost zero new energy output. As analysis of **Fig.13(c)** with bus_19_gen as the boundary revealed that these two typical operation scenarios encounter the situation wherein thermal power supply nodes cannot provide energy owing to resource shortages or failures under the high proportion of new energy integration scenario. Moreover, they are in the trough period of new energy output. While dispatching limited energy storage, more electricity must be dispatched across areas. In addition, in the scenario shown in **Fig.13(d)** where bus_113_gen serves as the boundary, there is a substantial thermal power output in the region with almost no new energy output, and a higher energy storage output is observed above the boundary. Therefore, the scenarios $OM_4$ and $OM_{11}$ involve the combined participation of the thermal power and energy storage. The analysis shows that both scenarios encounter a trough period of new energy output, which is seasonal and occurs because of the intermittent or fluctuating operation characteristics of new energy sources. However, they were independently extracted because scenario $OM_{11}$ involved the participation the adjustable loads in scheduling and operation.

***5.3 Analysis of efficiency and complexity***

To further demonstrate the advantages and practicality of DTSAs for extracting typical operating scenarios, this section uses the accumulated operational snapshot data of Area 3 as the dataset to compare the clustering results of the five methods: K-means, GMM, PCA_K-means, PCA_GMM and GAN_GMM. The historical operational snapshot data of Area 3 was selected because it represents a high-level stage in the evolution of the new power system. At this stage, new energy generation is the main source of power supply, and the complexity of grid operation necessitates higher performance requirements for typical operational scenario extraction methods that must coordinate various energy storage and flexible loads. **Table 2** compares the scores of the Silhouette Coefficient (SC), Akaike Information Criterion (AIC), and Bayesian Information Criterion (BIC) indicators for the best state obtained from 10 experiments performed using different methods on the same dataset.

SC is used as an evaluation index for DTSAs to extract typical operating scenarios, reflecting the cohesion within the same operating scenario cluster and the dispersion between different operating scenario clusters. The value range of SC is [-1, 1]. When it is close to +1, it means that similar operating snapshots are successfully aggregated, and the distinction between typical operating scenario clusters is obvious. When SC is a negative value, it indicates that a operating snapshot is assigned to the wrong operating scenario cluster. For each running snapshot s, SC is calculated as follows:

$$SC_i = \frac{b_i - a_i}{\max\{a_i, b_i\}} \tag{18}$$

where $a_i$ represents the average distance from running snapshot $s_i$ to other samples in the cluster to which it belongs. $b_i$ represents the minimum value among the average distances of running snapshot $s_i$ to all samples in other clusters. The $SC_i$ of all samples within $S$ is averaged to obtain the final index value.

AIC is a standard for evaluating the complexity of different typical operation scenario extraction models and the goodness-of-fit of the data. It is defined as follows:

$$AIC = 2k - 2\ln(L) \tag{19}$$

where $k$ is the number of parameters in the corresponding model and $L$ is the corresponding maximum likelihood function.

The BIC penalty considers the sample size, preventing the extraction model from pursuing excessive accuracy and resulting in excessive complexity for operational scenarios with a large number of samples. It is defined as follows:

$$BIC = k\ln(n) - 2\ln(L) \tag{20}$$

where $k$ and $L$ have the same meaning as in Equation (19). Further, $n$ is the sample size; the smaller the value, the better the comprehensive performance of the model.

Table 2: Performance comparison of different typical operation scenario extraction methods.

| | SC | AIC | BIC |
|---|---|---|---|
| K-means | 0.364 | 5018 | 5342 |
| GMM | 0.471 | 4986 | 5113 |
| PCA_K-means | 0.686 | 2165 | 2344 |
| PCA_GMM | 0.723 | 1383 | 1401 |
| GAN_GMM | 0.897 | 573 | 682 |
| DTSAs | 0.941 | 516 | 739 |

Upon analysis of the scores of various indicators presented in Table 2, it was found that k-means, being a hard clustering algorithm, struggled to converge and demonstrated poor clustering performance while dealing with high-dimensional nonlinear power grid operation data on a long-term scale. On the other hand, the GMM showed better performance than the k-means algorithm, with improvements in all three indicators. The primary reason for this is that the K-means algorithm is a heuristic algorithm that relies on Euclidean distance calculations, which can effectively measure low-dimensional data but fails in high-dimensional space due to the distortion of meaning. In contrast, the GMM algorithm is based on the likelihood of the probability density function and can measure distances better in high-dimensional space. Both PCA_K-means and PCA_GMM perform principal component analysis on the high-dimensional features of the power grid operation scenario, which reduces the complexity of model aggregation and significantly improves clustering performance. However, this method disrupts the integrity of the entire operating scenario, and there are certain bottlenecks in extracting the typical operating scenarios.

Compared with the above four methods, the performance indicators based on deep clustering methods (GAN_GMM, DTSAs) have been significantly improved. However, GAN has the problems of relatively unstable training process and mode collapse. Especially in the early stages of training, the generator will produce low-quality samples and require a lot of time to optimize and adjust training parameters. The superiority of DTSAs mainly lies in the fact that the operating snapshot image feature extraction link of this method is implemented by VGG13_GMP. Therefore, the introduction of excessive parameters did not decrease the values of the various indicators. Moreover, the key features of typical operating scenarios could be accurately extracted while ensuring the integrity of the operating scenarios. This also improves the within-cluster cohesion and between-cluster dispersion of the clustering results of the proposed method, and the SC index achieves its highest value of 0.941. The advantages of the AIC and BIC scores are mainly attributed to their comprehensive evaluation of the both model complexity and goodness-of-fit. However, the BIC places more emphasis on penalizing the sample size and model parameters, resulting in slightly higher scores than the AIC.

To summarize, after comparing and analyzing different methods for extracting typical operation scenarios, DTSAs were found to be suitable for adapting to changes in the penetration rate of new energy sources in regional power systems and effectively exploring typical operation scenarios and rules in power grid operational snapshots. The coordinated scheduling of generation, transmission, load, and storage modes should be a priority in the rapidly developing new power system. To cope with drastic changes in the system operation mode, it is important to reserve more typical operation scenarios in power grids with a high proportion of new energy penetration, which can assist dispatchers in quickly making dispatch decisions and ensure the economic, safe, and low-carbon operation of the system.

## 6 Conclusion

Owing to the rapid development of new power system, there is a higher demand for intraday dispatch plans to be adaptable to uncertainties in the multiple links of power supply and demand, as well as the complexity of the control targets. In this study, a method called DTSAs was introduced for extracting typical operating scenarios in power systems. DTSAs rely on a large amount of historical data and perform three key steps: encoding operating scenario sequences, extracting deep features from operating scenario images, and generating typical operating scenarios. This method captures the spatio-temporal characteristics of new power systems and extracts typical operating scenarios that align with the distribution of historical data.

The case studies conducted in this research, both qualitative and quantitative, demonstrated that the proposed DTSAs method outperforms traditional feature selection methods. Moreover, the experimental results revealed that our approach is robust and can extract typical operating scenarios for power systems with varying levels of new energy penetration. As a data-driven method, it enables dispatchers to gain a better understanding of power system operation experiences. In future work, it would be valuable to integrate typical operating scenario-based scheduling and optimization models for new power systems with expert knowledge in scheduling.

## Data Availability Statement

Research data are not shared.

## Conflict of Interest Statement

The authors declare no conflicts of interest.

## Acknowledgements

This work was supported by the Key R&D Project of Jilin Province (Grant No. 20230201067GX).